\documentclass[conference]{IEEEtran}
\usepackage[english]{babel}
\usepackage{amsmath}
\usepackage{amsfonts}
\usepackage[amsmath]{ntheorem}
\usepackage{graphicx}
\usepackage[colorlinks=true, allcolors=blue]{hyperref}
\usepackage[nolist]{acronym}
\usepackage{blkarray}
\usepackage{color}
\usepackage{amssymb}
\usepackage{graphicx}
\usepackage{tikz}
\usetikzlibrary{positioning}
\usetikzlibrary{calc,bending}
\usepackage{babel,blindtext}
\usepackage[capitalize]{cleveref}
\usepackage{algorithm}
\usepackage{algpseudocode}
\usepackage{bbm}
\usepackage[normalem]{ulem}
\usepackage{mathtools}
\usepackage{balance}

\acrodef{rl}[RL]{Reinforcement Learning}
\acrodef{ot}[OT]{Optimal Transport}
\acrodef{snr}[SNR]{Signal to Noise Ratio}
\acrodef{nns}[NN]{Neural Networks}

\newcommand{\enc}{\lambda}              
\newcommand{\dec}{\gamma}                                
\newcommand{\mt}{T}            
\newcommand{\mct}{\mathcal{T}}     
  
\newcommand{\q}{Q}                
                
\newcommand{\obs}{o}                     
\newcommand{\sem}{x}             
\newcommand{\act}{a}          
\newcommand{\sspace}{\mathcal{X}}    
\newcommand{\ospace}{\mathcal{O}}       
\newcommand{\aspace}{\mathcal{A}}                                 
\newcommand{\lang}{\ell}                                          
\newcommand{\satom}{P}                                            
\newcommand{\satomsource}[1]{P_{#1}^{\lang_s}}                    
\newcommand{\satomtarget}[1]{P_{#1}^{\lang_t}}                    
\newcommand{\infotransfer}[2]{\rho_{#1\to#2}}

\newtheorem{exmp}{Example}
\Crefname{exmp}{Example}{Examples}
\DeclareMathOperator*{\argmax}{arg\,max}

\newcommand\blfootnote[1]{
  \begingroup
  \renewcommand\thefootnote{}\footnote{#1}
  \addtocounter{footnote}{-1}
  \endgroup
}
\renewcommand{\baselinestretch}{0.93}
\usepackage{flushend}
\title{Semantic Language Mismatch Equalization}
\title{Latent Space Alignment for Semantic Channel Equalization}

\author{\IEEEauthorblockN{Tomás Hüttebräucker, Mohamed Sana, Emilio Calvanese Strinati}

\IEEEauthorblockA{CEA-Leti, Université Grenoble Alpes, F-38000 Grenoble, France\\
Email : \{tomas.huttebraucker; mohamed.sana; emilio.calvanese-strinati\}@cea.fr}}

\newcommand{\titleheader}{This work has been accepted for publication in 2024 International Conference on Machine Learning for Communication and Networking}
\makeatletter
\def\ps@IEEEtitlepagestyle{
\def\@oddhead{\mbox{}\scriptsize \titleheader \rightmark \hfil }
}
\makeatother

\begin{document}

\maketitle\blfootnote{The present work was supported by the EU Horizon 2020 Marie Skłodowska-Curie ITN Greenedge (GA No. 953775), by the France 2030 ANR program ``PEPR Networks of the Future" (ref. 22-PEFT-0010) and by the 6G-GOALS Project under the HORIZON program (no. 101139232).}  
\maketitle
\begin{abstract}
We relax the constraint of a shared language between agents in a semantic and goal-oriented communication system to explore the effect of language mismatch in distributed task solving. We propose a mathematical framework, which provides a modelling and a measure of the semantic distortion introduced in the communication when agents use distinct languages. We then propose a new approach to semantic channel equalization with proven effectiveness through numerical evaluations.
\end{abstract}

\section{Introduction}
Semantic and goal--oriented communications \cite{calvanese20216g} \cite{kountoris2021} emerge as a paradigm shift from Shannon's classical view of communication. By transmitting only the meaning and task--relevant information extracted from the data, semantic protocols (languages) reduce network resource consumption, thereby increasing its capacity to host new services. The design of a semantic language between intelligent agents is an active area of research, showing promising results \cite{tung2021effective} \cite{xie2021deep}. Many works in the literature assume that the language is agreed upon and shared by the agents. However, in many scenarios, the communication context and task are constantly changing; therefore, the language should adapt to these changes. Continuously learning a language between multiple agents is a resource-hungry procedure, which in a limited energy and bandwidth network is infeasible. In this work, we relax the shared language constraint, allowing agents to use distinct languages. We explore the effects of this \textit{language mismatch} on a task performance. Moreover, we mathematically model the mismatch as systematic errors caused by the \textit{semantic channel} and we propose, as it is also usual for the systematic errors caused by the physical channel, an equalization algorithm.

\section{System Model}
\begin{figure}[!t]
    \centering
    \includegraphics[width=\linewidth]{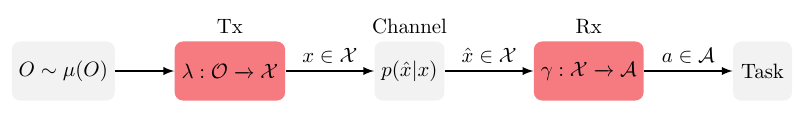}
    \vspace{-0.5cm}
    \caption{System model.}
    \label{fig:system_model}
\end{figure}

We consider the system model of \cref{fig:system_model}. An agent at the transmitter, observes the world and uses a \emph{language generator} $\enc$ to extract and encode the underlying information into a semantic representation. Such representation is then conveyed through a noisy channel to the receiver, where it gets interpreted by another agent using a \emph{language interpreter} $\dec$, which maps it to an action aiming to complete a task. In this context, a good language representation is instrumental in structuring the semantic information exchanged between agents to accomplish the task. We define a language $\lang=(\enc, \dec,\mu, \ospace, \sspace, \aspace)$ as a tuple formed by an observation space $\ospace$, a semantic space $\sspace$, an action space $\aspace$ jointly with a distribution $\mu$ over $\ospace$ and a language generator $\enc:\ospace\xrightarrow{}\sspace$ and a language interpreter $\dec:\sspace\xrightarrow{}\aspace$, possibly stochastic. A language generator $\enc$ endows observation $o\in\ospace$ with a semantic symbol $x\in\sspace$, which is mapped to action $a=\dec(x) \in \aspace$ by the language interpreter. A good language defines a \textit{partition} over the semantic space. A partition $\satom=\{\satom_1,\satom_2,\dots,\satom_N\}$ of the semantic space $\sspace$ is a collection of measurable sets - \emph{called atoms of the partition} - that cover $\sspace$. Each atom is associated with a semantic meaning. For example, in an image classification task, an atom might be associated to a given class, or to a given characteristic of the image (color, shape, etc.). 
In our study, both the language generator and interpreter are artificially modeled by \ac{nns}. 
Ideally, a communication between two distinct agents assumes a shared language. In other words, both language interpreter and generator should be compatible and the result of an end-to-end learning procedure. 
However, in practical scenarios, agents may use distinct languages, i.e. different logic for extracting and representing semantic information. In such case, communication is prone to semantic-errors due to language mismatch. Consider the following example.
\begin{exmp}[The scout and the treasure]\label{exmpl:scenario}
    A treasure and a scout are randomly placed in a discrete grid. At each time step, the encoder observes the state of the environment and transmits a semantic symbol $\sem\in\sspace=\mathbbm{R^2}$ through a Gaussian noisy channel. The decoder (scout) receives the noisy symbol and takes one of four actions  $\aspace=\{\text{right, down, left, up}\}$. The episode ends when the scout reaches the treasure or the maximum number of steps is attained.
\end{exmp}
\begin{figure}[!ht]
    \centering
    \includegraphics[width=0.85\linewidth]{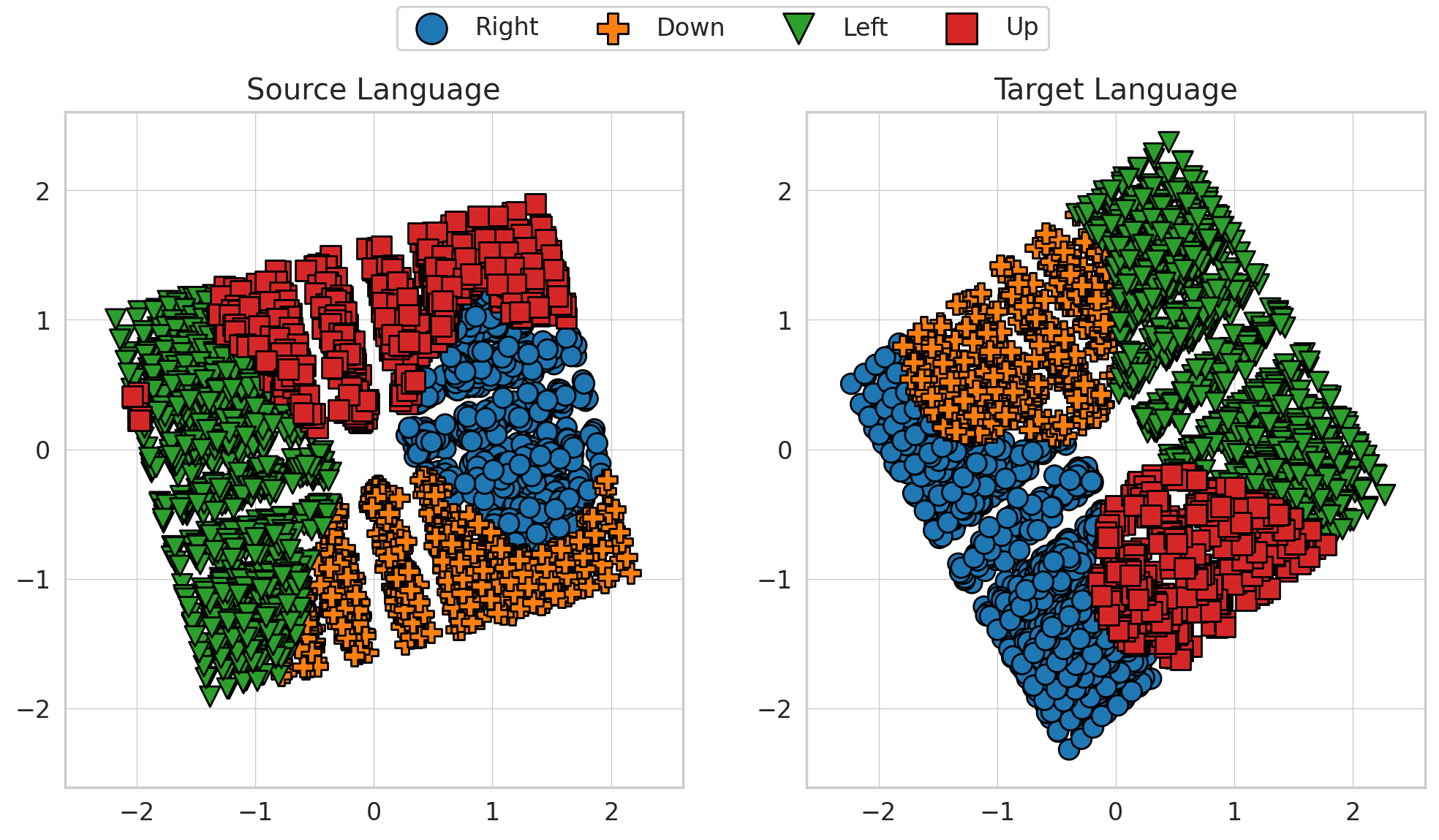}  
    \caption{Two languages solving the same task, learned under the same conditions using reinforcement learning, but with different semantic representation.}
    \label{fig:language_mapping}
\end{figure}
In \cref{fig:language_mapping} the partition of the semantic space for two different languages (denoted source and target language) that followed independent learning procedures is shown. As we can see, the two languages partition the semantic space differently. Semantic symbols transmitted by the source language generator will be incorrectly understood by the target language interpreter unless they fall in the corresponding target atom. Indeed, if each source atom was contained in its corresponding target atom, there will be no errors due to the language mismatch. On the contrary, if the intersection between each corresponding pair of source and target atoms is empty, all symbols will be misinterpreted. 

Therefore, \textit{the language mismatch can be modeled as a misalignment of the semantic space partitions}.

\section{Latent space alignment}

Let denote with $\lang_s$ and $\lang_t$ the source and target language used at the transmitter and receiver respectively. To align the latent spaces of the source and target, we propose to operate a codebook of transformations over the semantic space. Specifically, for each pair of source and target atoms $(\satomsource{i}, \satom^{\lang_t}_j)$, we look for transformation $\mt:\sspace\xrightarrow{}\sspace$, that maximizes the following information transfer metric \cite{sana2023semantic}: 
\begin{equation}\label{eq:information_transfer}
\begin{split}
    \infotransfer{i}{j}(\mt) =   \frac{\mu_{\mt\enc_s}\left(\mt\left (\satomsource{i} \right )\cap \satom^{\lang_t}_j\right)}{\mu_{\mt\enc_s}\left(\satomsource{i}\right)},
\end{split}
\end{equation}
where $\mu_{\mt\enc_s}$ is a probability measure over the semantic space (defined as $\mu_{\mt\enc}(\sem)=\sum_{\obs\in\ospace} \mt\circ\enc(\sem|\obs) \mu(\obs)$ for a stochastic generator $\enc$). Intuitively, $\infotransfer{i}{j}(\mt)$ measures how much of the volume of $\satomsource{i}$ is transported into $\satom^{\lang_t}_j$ by the transformation $T$. Therefore, it is a measure of how well $\mt$ transforms the meaning of atom $\satomsource{i}$ so that it is interpreted as $\satomtarget{j}$ by the receiver. To learn low-complexity codebook, we assume linear transformations, which we learn using \ac{ot} \cite{perrot2016mapping} for all pairs of atoms.

\section{Codebook operation policy}
To operate the codebook, we propose two different policies as in \cite{huttebraucker2023pragmatic}: 

\noindent
\textbf{Semantic equalization policy.}
The semantic equalization policy $\pi_\text{sem}$ operates the codebook to compensate semantic mismatch (i.e. to perfectly align language partitions). For each observation $\obs$, the transformation $\mt$ maps the semantic representation $\enc(\obs)$ into its corresponding semantic atom for the target language. We obtain $\pi_\text{sem}$ by solving:

\begin{equation*}\label{eq:sem_risk}
    \pi_\text{sem} = 
    \argmax_{\mt\in \mct}\Bigg[ \sum_{i\in J_s} \mu_{\enc_s}\left (\satomsource{i}|\obs \right)  \sum_{j\in\kappa(i)} \infotransfer{i}{j}(\mt) \Bigg ]
\end{equation*}
where $\kappa:J_s\xrightarrow{}J_t$ captures the correspondence between source and target atoms, $J_s$ and $J_t$ denote the index functions of source and target;  $\mu_{\enc_s}\left (\satomsource{i}|\obs \right)$ is the probability that the semantic symbol $x=\enc_s(\obs)$ belongs to the semantic atom $\satomsource{i}$. 

\noindent
\textbf{Effectiveness equalization policy.}
Perfect semantic alignment is not necessary for effective task solving. Indeed, even if the semantic meaning is incorrectly interpreted, the action to which this interpretation leads might still correct. The effectiveness equalization policy $\pi_\text{eff}$ focuses on the end-goal decision rather than the intermediate meaning:

\begin{equation*}\label{eq:eff_risk}
    \pi_\text{eff} =  \argmax_{\mt\in\mct}\Bigg[ \sum_{i\in J_s} \mu_{\enc_s}\left (\satomsource{i}|\obs \right) \sum_{j\in J_t} \infotransfer{i}{j}(\mt) \q(\act_j,\obs) \Bigg ]
\end{equation*}
where $\q$ is the \ac{rl} Q-value function. Here, $\q(\act,\obs)$ indicates how optimal is action $\act$ given that the observation of the environment is $\obs$.

\section{Results and Conclusions}
\begin{figure}
    \centering
    \includegraphics[width=.7\linewidth]{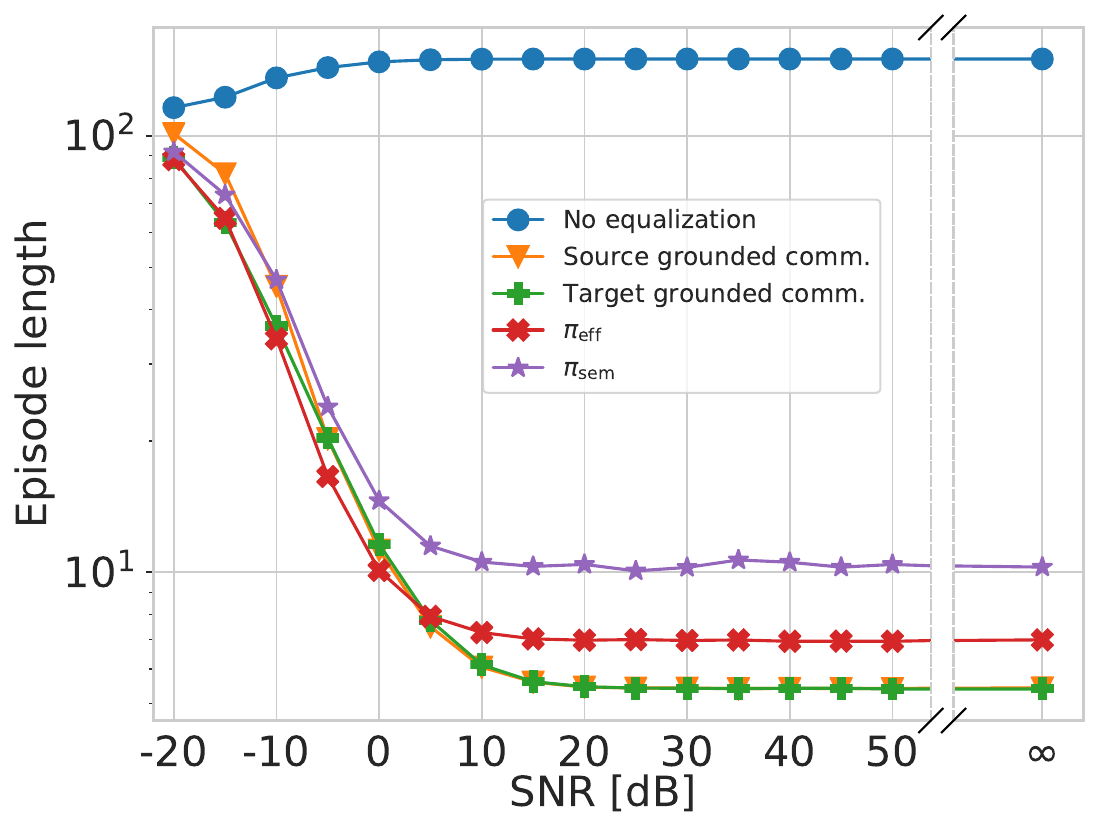}
    \caption{Average episode length (lower is better) for the different communication strategies with varying \ac{snr} for a stochastic decoder.}
    \label{fig:performance_stochastic}
\end{figure}

Considering the aforementioned \cref{exmpl:scenario}, we explore the language mismatch effects on the task performance and how effective the proposed equalization algorithms are. In \cref{fig:performance_stochastic} the performances of the proposed equalization methods are compared to the no equalization approach and the source and target language baselines. As it is shown, the equalization methods are effective at solving the language mismatch and attain close performance to the no mismatch cases. We observe as well that the $\pi_\text{eff}$ policy performs better than its semantic counterpart $\pi_\text{sem}$. This can be explained by the goal--oriented nature of the effectiveness policy, which prioritizes task performance (measured by $\q$) rather than perfect semantic alignment.

Future work will explore new approaches for defining semantic space atoms without resorting directly to actions, which limit the expressivity of the semantic space.

\balance
\bibliographystyle{ieeetr}
\bibliography{bib}

\end{document}